\documentclass[lettersize,journal]{IEEEtran}
\usepackage{amsmath,amsfonts}
\usepackage{algorithmic}
\usepackage{algorithm}

\usepackage{array}
\usepackage[caption=false,font=normalsize,labelfont=sf,textfont=sf]{subfig}
\usepackage{textcomp}
\usepackage{stfloats}
\usepackage{url}
\usepackage{verbatim}
\usepackage{graphicx}
\usepackage{cite}
\usepackage{booktabs}
\usepackage{multirow}
\usepackage{amssymb}
\usepackage{amsmath}
\usepackage{color}
\usepackage{ragged2e}
\usepackage[colorlinks=black,linkcolor=black]{hyperref}

\begin{document}
	
\title{Masked Graph Learning with Recurrent Alignment for Multimodal  Emotion Recognition  in Conversation}
	
\author{Tao~Meng, Fuchen~Zhang, Yuntao~Shou, Hongen Shao, Wei~Ai and~Keqin~Li,~\IEEEmembership{Fellow,~IEEE}
    \thanks{Corresponding Author: Wei Ai~(aiwei@hnu.edu.cn)}
	\thanks{This work is supported by National Natural Science Foundation of China (Grant No. 69189338, Grant No. 62372478), the Research Foundation of Education Bureau of Hunan Province of China (Grant No. 22B0275), and the Changsha Natural Science Foundation (Grant No. kq2202294).}
	\IEEEcompsocitemizethanks{\IEEEcompsocthanksitem T. Meng, F. Zhang, Y. Shou H. Shao and W. Ai are with College of Computer and Mathematics, Central South University of Forestry and Technology, Changsha, Hunan 410004, China. (mengtao@hnu.edu.cn, fuchen.zhang@csuft.edu.cn, yuntaoshou@csuft.edu.cn, hongen.shao@csuft.edu.cn, aiwei@hnu.edu.cn)
	\IEEEcompsocthanksitem K. Li is with the Department of Computer Science, State University of New York, New Paltz, New York 12561, USA. (lik@newpaltz.edu)}}

	
	
\maketitle

\begin{abstract}
Since Multimodal Emotion Recognition in Conversation (MERC) can be applied to public opinion monitoring, intelligent dialogue robots, and other fields, it has received extensive research attention in recent years. Unlike traditional unimodal emotion recognition, MERC can fuse complementary semantic information between multiple modalities (e.g., text, audio, and vision) to improve emotion recognition. However, previous work ignored the inter-modal alignment process and the intra-modal noise information before multimodal fusion but directly fuses multimodal features, which will hinder the model for representation learning. In this study, we have developed a novel approach called Masked Graph Learning with Recursive Alignment (MGLRA) to tackle this problem, which uses a recurrent iterative module with memory to align multimodal features, and then uses the masked GCN for multimodal feature fusion. First, we employ LSTM to capture contextual information and use a graph attention-filtering mechanism to eliminate noise effectively within the modality. Second, we build a recurrent iteration module with a memory function, which can use communication between different modalities to eliminate the gap between modalities and achieve the preliminary alignment of features between modalities. Then, a cross-modal multi-head attention mechanism is introduced to achieve feature alignment between modalities and construct a masked GCN for multimodal feature fusion, which can perform random mask reconstruction on the nodes in the graph to obtain better node feature representation. Finally, we utilize a multilayer perceptron (MLP) for emotion recognition. Extensive experiments on two benchmark datasets (i.e., IEMOCAP and MELD) demonstrate that {MGLRA} outperforms state-of-the-art methods.
\end{abstract}
	
\section{Introduction}
\label{sec:introduction}
\IEEEPARstart{E}motions {affect every aspect of our lives through thoughts or actions, and conversation is the primary way to express them \cite{4, shou2022conversational, shou2022object, ying2021prediction, meng2024multi, shou2023low, shou2023graph}. Therefore, it is crucial to understand emotions in conversation accurately, and the results can be widely used in fields such as intelligent dialogue\cite{5} and intelligent recommendations \cite{2}. However, in actual dialogue scenes, the emotions expressed by the speaker are not only related to the content of the speech but also closely related to his tone and expression. {Multimodal Emotion Recognition in Conversation (MERC) task aims to use the utterance (text, audio) and visual (expression) information in the conversation to identify the speaker's emotion.} Compared with traditional unimodal emotion recognition in conversation, MERC can improve the instability of emotion analysis by fusing richer multimodal semantic information \cite{7,8, shou2023czl, shou2023comprehensive, ai2023gcn, meng2023deep, shou2023adversarial, SHOU2024102590}. Therefore, the key to advancing MERC lies in the effective alignment and fusion of text, audio, and visual information to achieve a collaborative understanding of cross-modal emotional semantics. \cite{shou2023masked, shou2024revisiting, shou2024efficient, ai2023two, meng2024revisiting}}

{In response to the above challenges, many researchers have made significant efforts in the field of conversational emotion recognition. For instance, \textit{Liu et al.}\cite{6} first used a convolutional neural network (CNN) to learn local features of speech signals, then used a recurrent neural network (RNN) to capture long sequence features, and finally fused the two types of features to achieve emotion recognition.} \textit{Lian et al.} \cite{9} propose a transformer-based dialogue emotion recognition model called CTNet, which can adaptively learn and capture important emotional features from input dialogues. Due to the excellent performance of graph neural networks (GNN) in relational modeling,  \textit{Ghosal et al.}\cite{7} converted the conversation history into a graph data structure and effectively extracted the emotional features in the conversation history through GNN. This method can not only be used for emotion recognition tasks but also can be applied to other dialogue-related tasks. \textit{Hu et al.} \cite{8} {use} GNN to model speaker-to-speaker relationships, effectively exploiting multimodal dependencies and speaker information. {\textit{Yuan et al.} \cite{yuan2023rba} used the relational bilevel GNN to model MERC, which reduced the redundancy of node information and improved the capture of long-distance dependencies.}

\begin{figure*}[htbp]
	\centering
	\includegraphics[width=1.0\linewidth,scale=1.00]{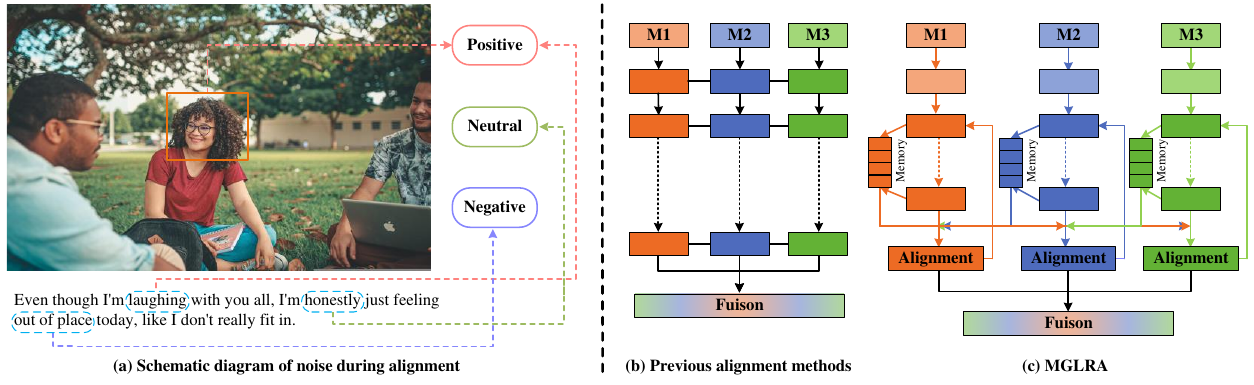}
	\caption{{An example to illustrate the importance of alignment before performing multi-modal fusion and the difference from existing methods. (a) The example demonstrates the first type of noise in multimodal emotion recognition. Red and blue represent visual information and textual information, respectively. (b) Previous alignment methods for MERC. (c) Our alignment method MGLRA.}}
	\label{fig_1}
\end{figure*}
	
However, only considering multimodal fusion is not complete enough for MERC. The alignment of semantic features before multimodal fusion is also a difficult challenge for MERC, which affects the fusion performance to some extent. Alignment is often used to unify disparate data from multiple modalities \cite{10,11}. {At the same time, noise reduction processing is indispensable during the alignment process. There are usually two types of noise. (1) As shown in Fig. \ref{fig_1} {(a)}, features with different granularities may mean inconsistent emotional polarity. Character visual angles represent positive emotions, while different words represent neutral or negative emotions. (2) The original features extracted from the corresponding single modality using different pre-trained models may contain some missing, redundant, or even wrong information. Previous researchers have done a lot of work on this issue.} For example, \textit{Chen et al.}\cite{12} proposed a gated multimodal embedding LSTM, which can filter noise information while processing noisy modality data and achieve finer fusion between input modalities. \textit{Xu et al.} \cite{xu2019learning} use the attention mechanism to use an adaptive alignment strategy in the alignment layer, which can automatically learn alignment weights in the process of frame and word alignment in the time domain. \textit{Xue et al.} \cite{xue2022multi} proposed a multi-level attention map network to filter intra-modal or inter-modal noise to achieve fine-grained feature alignment. However, {as shown in Fig. \ref{fig_1} (b), these methods have the following limitations: (1) The alignment process is often completed in one go, lacking an iterative alignment process, resulting in the model being unable to complete fine-grained alignment. (2) These methods do not allow the model to observe representations extracted from other modalities and realign them during the alignment process and do not consider contextual dialogue relationships during the alignment process, resulting in poor performance when dealing with the first type of noise.}
	
To align and fuse semantic information from multiple modalities, we carry out a Masked Graph Learning with Recurrent Alignment for Multimodal  Emotion Recognition  in Conversation (MGLRA), which uses an iterative alignment mechanism to strengthen the modalities' consistency gradually, {as shown in Fig. \ref{fig_1} (c).} First, MGLRA uses different feature encoders to represent modality-specific features. Second, we employ LSTM to capture contextual information and use a graph attention-filtering mechanism to eliminate noise effectively within the modality. Third, MGLRA employs a novel feature alignment method based on recursive memory-augmented cross-modal attention, which iteratively refines and aligns unimodal features by observing memory modules from other modalities. Fourth, we introduce a variational GCN-based fusion method to fuse unimodal features to produce a robust multimodal representation. Finally, multimodal representation is directly used in MERC to generate emotion classification. The contributions of our work are summarized as follows:
\begin{itemize}
	\item[$\bullet$] We propose a novel Masked Graph Learning with a Recurrent Alignment (MGLRA) model to refine unimodal representations of semantic information from multiple modalities. {MGLRA uses a memory mechanism to iteratively align semantic information from multiple modalities, making it more robust in noisy scenes and scenes lacking modal information.}
		
	\item[$\bullet$] We introduce a cross-modal multi-head attention mechanism to explore interactive semantic information among multiple modalities and expand the receptive field of contextual information.
		
	\item[$\bullet$] We utilize a simple and effective GCN with a random masking mechanism to fuse complementary semantic information among multiple modalities without introducing extra computation.
		
	\item[$\bullet$] Extensive experiments are conducted on two widely used benchmark datasets (i.e., IEMOCAP and MELD) to verify the effectiveness of the proposed model. The experimental results show that the model proposed in this paper is superior to the existing comparison algorithms {regarding accuracy} and F1-score.
\end{itemize}
	
We organize the subsequent {sections} of this paper as follows: Section \ref{sec:relatedwork} presents related work on MERC; Section \ref{sec:preliminary} introduces the model's feature extraction part and the mathematical part's definition and interpretation; Section \ref{sec:methodology} {describes} the details of our proposed model; Sections \ref{sec:experiments} and \ref{sec:resultsdiscussion} present and analyze our experimental {results, and} the conclusion is arranged in Section \ref{sec:conclusion}.

\section{Related work}
\label{sec:relatedwork}
In this section, we mainly introduce the research in emotion recognition, {the technology of alignment mechanism related to our research}, and the latest related work of GCN.
	
\subsection{Multimodal Emotion Recognition in Conversation}
{Conversation analysis and speaker relationship modeling are crucial in emotion recognition tasks. As demonstrated in \cite{14} and \cite{15}, previous studies have deeply explored the relationship between emotion and social relevance in conversations and highlighted the dynamic emotional issues arising in human interactions. These interactions form complex interrelationships that require dynamic consideration in model design. To this end, DialogueRNN \cite{16} adopts a recurrent neural network (RNN) and an attention mechanism to automatically learn long-term dependencies and dynamic interaction patterns between speakers, {which shows} excellent performance. DialogueGCN \cite{7} transfers and aggregates information on the nodes and edges of the graph structure through graph convolutional neural network (GCN), more effectively handles long-term dependencies and multi-round dialogue interactions, and better captures the global structure and context of dialogues. Inspired by the high-performance models in multi-modal community tasks \cite{18,19,20,21}, many current models focus on the fusion process to solve the multi-modal emotion recognition (MERC) task. For example, LR-GCN \cite{17} introduces a latent relationship representation learning mechanism to better represent the interactive relationships between nodes and edges by learning the latent connections between nodes, thereby improving performance.}
	
{However, these methods mainly focus on the fusion process of context information and speaker information to improve the performance of the fusion process but ignore the semantic information alignment process before fusion. This directly limits the effectiveness of the model during fusion.}
	
\subsection{Alignment Mechanism}
The alignment mechanism matches the semantic features from different modalities so that the emotional expressions of other modalities are consistent. Multimodal emotion recognition mainly uses context-order-based and enforced word-level alignments.
	
The alignment of the contextual order relationship can be used to compare the similarity or difference between two sentences while aligning the relationship between words and {frames to} achieve the fusion between modalities. \textit{Liu et al.} \cite{22} used the attention mechanism to establish the alignment between vision and audio modalities, and corresponding elements can be found in the two modalities. This correspondence is called bi-directional attention alignment. \textit{Li et al.} \cite{23} proposed an Inter-modality Excitement Encoder (IMEE), which can learn the refined excitability between modalities, such as vision and audio modalities. \textit{Chen et al.} \cite{24} introduced a cross-modal time consistency module, which used a Bi-LSTM model to learn the time dependence between vision and audio modalities, ensuring that the emotional prediction results of the two modalities at the same time step are consistent.
	
Enforced word-level alignment in tasks with multiple modal inputs by providing information in the decoder corresponding to words or regions in different modalities; forced alignment between other modalities and output sequences is achieved. \textit{Gu et al.} \cite{25} proposed a method for multimodal emotion analysis using a hierarchical attention mechanism and word-level alignment. \textit{Romanian et al.} \cite{26} implemented a technique for detecting depression using word-level multimodal fusion, which mainly relies on time-based recursive methods to achieve word-level modal alignment, but word-level processing also brings high computational complexity, which is something to consider.
	
Despite the success of the above methods, they are all local alignment mechanisms, ignoring the interaction of global context information, which leads to limited context awareness of the model.
	
\subsection{Graph Convolutional Network}
{The rise of graph neural networks (GNN) has attracted researchers over the past few years. It has achieved remarkable success in research areas such as semantic segmentation, object detection, and knowledge graphs\cite{27,28}. The graph convolutional network (GCN) proposed by \textit{Kipf et al.} \cite{29} is central to this success. The technology is similar to traditional convolutional neural networks (CNN), using convolutions to pass information through the network and capture comprehensive data set features. Its efficiency stems from leveraging unlabeled data for model augmentation, achieved using a simple similarity matrix. \textit{Veljkovic et al.} \cite{30} recently developed a graph attention network (GAT). This model innovatively uses the attention mechanism to dynamically understand the graph structure, thereby achieving accurate modeling and reasoning of complex relationships in the graph. This enables superior performance for tasks such as node classification and inference. The progress of GCN has achieved significant breakthroughs, motivating us to apply GCN to address challenges in MERC.}

{Although the effectiveness of deep learning methods has been proven, the MERC task still needs to improve, such as semantic consistency between different modalities and the complexity of effectively fusing multi-modal features. To address these challenges, our paper introduces the MGLRA framework. This novel framework adopts a recurrent alignment strategy enhanced by a memory component to ensure semantic alignment across modalities before fusion. Furthermore, we utilize a multi-head attention mechanism to explore the relationships between modules in detail and combine it with a computationally efficient GCN for effective fusion.}

\section{Preliminary}
\label{sec:preliminary}
In this section, we introduce various details of our work from a data flow perspective and how we extract multimodal features of all utterances from the dataset.
	
\subsection{Multimodal Feature Extraction}The reason for unimodal-specific feature extraction methods is twofold. First, since each modality has its unique semantic features, it is best to use its special feature extractor for each modality to capture salient representations adequately. Second, this unimodal feature extraction method allows state-of-the-art feature extractors to obtain better unimodal semantic information for transfer learning. The feature extraction methods and different processing methods of all modalities used are as {follows:}
	
\subsubsection{Text Feature Extraction}
Emotion keywords in text content play a crucial role in emotion recognition, so extracting rich lexical features from text content is a fundamental challenge in multimodal emotion recognition. In addition, the contextual semantic information composed of sentences as the basic unit also provides many guiding clues for emotion recognition. In multimodal emotion recognition research, RoBERTa often captures local semantic and global contextual features. In this paper, following previous work \cite{16, 7, 34}, we also adopt {the RoBERTa} model to extract and represent text features. The final processed text features are represented as $x_t$, and  $x_t \in \mathbb{R}^{d_t}$, $d_t=100$. The sequence of text features is denoted by ${X_t}$.

\subsubsection{Audio Feature Extraction}In determining the speaker's emotional state, audio features play a crucial role in information. So we use openSMILE to extract audio features following previous work \cite{16, 7, 34}. {It} is a highly open-source software for extracting audio features, mainly used in emotion recognition, emotion computing, music information processing, etc. It can extract many vectors in audio, including MFCC, frame intensity, frame energy, pitch, etc. The final processed audio features are represented as $x_a$, and  $x_a \in \mathbb{R}^{d_a}$, $d_a=100$. The audio feature sequence is denoted by ${X_a}$.
	
\subsubsection{Vision Feature Extraction}
Since the facial expression features can best reflect the emotional changes at a particular moment, we use 3D-CNN to extract the expressive facial features of the interlocutor to enhance the extraction of unimodal features of the vision to pursue better multimodal fusion effect following previous work \cite{16, 7, 34}. In addition to extracting the details of relevant features from many key image frames, 3D-CNN can extract spatiotemporal features jumping across multiple image frames. Doing so makes it easy to identify critical emotional states, such as smiling or depression. The final processed vision features are represented as $x_v$, and  $x_v \in \mathbb{R}^{d_v}$, $d_v=512$. The vision feature sequence is denoted by  ${X_v}$.
	
\subsection{Problem Definition}
Given a dataset $\mathit{S}$ of multimodal dialogues with multiple characters, through our designed preprocessing process, we get input features $X_{m}^{r}=\left(x_{(m, 1)}, x_{(m, 2)}, \ldots, x_{\left(m, l_{m}^{r}\right)}\right), m \in\{a, v, t\}$, $a$, $v$, $t$ represent audio, vision, and text, respectively. Here, $\mathit{r}$ represents the original feature, and $l_m^r$ means the sequence length. Given these sequences $X_{m}$, the final task is to determine a deep fusion network $F(X_m)$ such that the output $\hat{y}_{m}$ is getting closer and closer to the target $y_m$, {which can be achieved by minimizing the loss function.} The loss function of the model is defined as shown in Eq. (1):

\begin{equation}
	\min _{F} \frac{1}{b} \sum_{m=1}^{b} \mathcal{L}\left(\hat{y}_{m}=F\left(X_{m}\right), y_{m}\right)
\end{equation}
where $b$ represents the batch size, $y_m$ is the true emotion of the utterance and $\hat{y}_{m}$ is the predicted emotion of the model.

\section{Methodology}
\label{sec:methodology}
\begin{figure*}[htbp]
	\centering
	\includegraphics[width=1.0\linewidth]{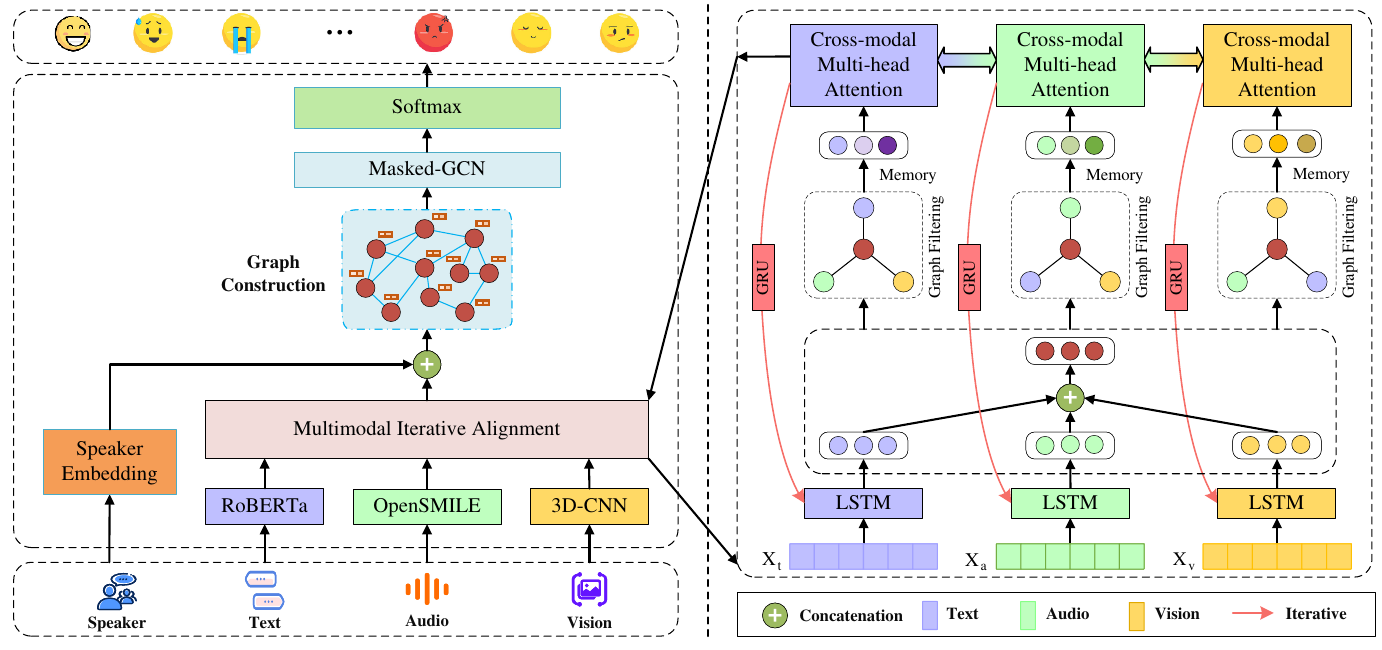}
	\caption{We propose the architecture of MGLRA. In the preprocessing stage, we use different feature extractors for the structural features of different modality data. In the multimodal feature alignment stage, we use a graph filtering mechanism for noise reduction and propose an alignment architecture with a memory iteration mechanism to enhance semantic features. Moreover, the speaker's information is incorporated into the construction process of the graph. Then the masked GCN is used to fuse the semantics to achieve the final emotion label classification.}
	\label{fig_2}
\end{figure*}
	
This section proposes a novel Masked Graph Learning with Recurrent Alignment (MGLRA) to improve emotion classification performance for multimodal emotion recognition in conversation. Fig. \ref{fig_2} shows the overall architecture of {MGLRA. This method} consists of the following parts: data preprocessing, multimodal iterative alignment, multimodal fusion with masked GCN, and emotion classifier. (1) Data preprocessing: Due to the different data structures of other modalities, we use RoBERTa, openSMILE, and 3D-CNN to extract text, audio, and visual features in the data preprocessing stage. (2) Multimodal iterative alignment: First, a graph attention filtering mechanism is developed in the multimodal feature alignment stage to adequately filter redundant noise in multimodal {features.} Second, to enhance the expressive ability of the original multimodal features, the memory-based recursive feature alignment (MRFA) was created, and this module was used to gradually realize the preliminary alignment of the three modalities using the memory iteration mechanism. Third, we develop cross-modal multi-head attention to discover shared information and complementary relationships between modalities to understand and express emotions better. (3) Multimodal fusion with masked GCN: For the fusion problem of multimodal emotion recognition in conversation, a simple and effective masked GCN is used for multimodal feature fusion, which achieves good performance without bringing more parameters. (4) Emotion classifier: We used MLP for emotion classification in the emotion classification stage.
	
\subsection{Multimodal Iterative Alignment}
A multimodal iterative alignment module is designed to improve the fusion effect of multimodal features, which provides aligned and robust unimodal representations for downstream fusion tasks. Specifically, due to the gap in the semantic information between the three modalities and the different peaks of their data distributions, it is difficult for the model to capture the complementary semantic information among the three modalities. Therefore, we cyclically align unimodal features before downstream task fusion to produce a significant unimodal representation. This unimodal feature alignment mechanism is also reflected in the multisensory cognitive system of animals. Technically, the multimodal iterative alignment module includes a graph attention filtering mechanism, a memory-based recursive feature alignment method, and cross-modal multi-head attention. Specifically, we first exclude redundant or wrong information within or between modalities through a graph attention filtering mechanism. Then, the memory-based recursive feature alignment is used to achieve the preliminary alignment of features between modalities. Finally, the final alignment of inter-modal features is achieved using cross-modal multi-head attention.
	
\subsubsection{Graph Attention Filtering Mechanism}
Inspired by the Multi-Level Attention Graph Network (MAMN)\cite{xue2022multi}, we use a graph attention filtering mechanism to filter out some noise in raw multimodal features, which may contain wrong, redundant, or missing information. The difference is that we assign higher weights to the more critical multimodal features rather than multi-granular features. After such a process, the representations of all multimodal features are re-optimized. Correspondingly, the weighted average based on the attention mechanism relatively emphasizes or weakens the role of a modality. This can lead to more accurate emotional judgments.
	
The graph structure must be constructed before utilizing the graph attention filtering mechanism for intra-modal or inter-modal noise reduction. For the input text, visual, and audio modality features, we first feed them into a long short-term memory network (LSTM) to extract contextual semantic information. The formula for LSTM is defined as follows:
\begin{equation}
	X_t^c,X_a^c,X_v^c =\text{{LSTM}} \left( {X_t^r,X_a^r,X_v^r} \right)
\end{equation}
where $r$ represents the original feature, and $c$ represents features with contextual information.
	
{Then,} we construct a graph structure for the captured multi-modal context information. The specific graph construction process is as follows. As shown in Fig. \ref{fig_2}, the central node of the graph is a multimodal feature node, represented by brown, and its features are generated by connecting text, audio, and visual features. The first-order neighbors of the central node are unimodal feature nodes, and purple, green, and yellow represent text, speech, and visual feature nodes, respectively. Moreover, the edges represent the relationships between each unimodality and multimodality, such as text-multimodal, audio-multimodal, and visual-multimodal relationships. Finally, the graph attention filtering mechanism filters noise or redundant information within and between modalities by learning information about nodes and edges and assigning different weights to different nodes.
	
The input of the graph attention filtering mechanism is the edge relation category matrix $C\in R^{{3}\times T}$ and the node eigenvalue matrix $V\in R^{{4}\times P}$, where $T$ is the dimension of each relation type embedding, and $P$ is the dimension of each eigenvalue. The relation category $C$ contains three basic semantic features: text-multimodal, audio-multimodal, and visual-multimodal levels. The eigenvalue matrix $V$ contains four kinds of semantic features: multimodal features $X_m^c$, text features $X_t^c$, audio features $X_a^c$, and visual features $X_v^c$.
	
In order to filter the noise or redundant information of each node, the relationship degree between feature node pairs is represented by $c_{ijk}$, which is defined in Eq. (3):
\begin{equation}
	c_{ijk}=W_{ijk}\left[V_i\left.||V_j\right.||C_k\right]
\end{equation}
where $W_{ijk}$ represents the linear transformation matrix obtained by learning in the network, $\parallel$ represents cascade operation. In particular, $V_i$, $V_j$, represents the $i$-th row and $j$-th column of the value matrix, and $C_k$ represents the $k$-th row of the edge relation category matrix.

The purpose of filtering and enhancing the original semantic features is achieved by using the attention mechanism to assign different weights to different multimodal feature vectors. The attention weight calculation formula is as follows:
\begin{equation}
	\alpha_{ijk}=\frac{\exp{\left(c_{ijk}\right)}}{\sum_{q\in Q_i}{\sum_{r\in R_{iq}}\ \exp{\left(c_{iqr}\right)}}}
\end{equation}
where $Q_i$ represents the neighbors of feature node $v_i$, and $R_{iq}$ represents the relationship type between feature node pair $v_i$ and $v_q$. For the feature node $v_i$, its filtering feature is the sum of each pair of representations weighted by their attention weights:
\begin{equation}
	h_i=\sum_{j\in Q_i}{\sum_{k\in R_{ij}}\ \alpha_{ijk}}c_{ijk}
\end{equation}
	
Each feature vector is updated through all the above computation steps, resulting in three new unimodal features $X_a^h, X_v^h, X_t^h$. Furthermore, the noise in each mode has been reduced.
	
\subsubsection{Memory-based Recursive Feature Alignment (MRFA)}
\begin{figure}[htbp]
	\centering
	\includegraphics[width=1.0\linewidth,scale=1.00]{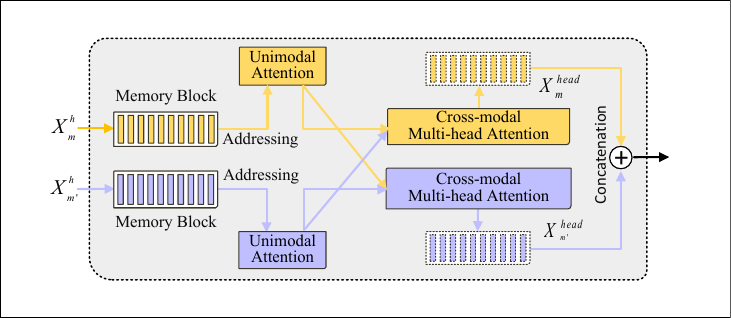}
	\caption{{Detailed pipeline for aligning multimodal data using MRFA and cross-modal multi-head attention. First, each modality has a corresponding memory block for information storage. Then, a single-modal attention mechanism is used to extract intra-modal information. Finally, cross-modal multi-head attention is used to achieve multi-modal feature fusion. Here we use two modes as examples, and the three modes in the paper cross each other in pairs.}}
	\label{fig_3}
\end{figure}

{A specific flowchart illustrating how we utilize MRFA and crossmodal multi-head attention for alignment is depicted in Fig. \ref{fig_3}.} In MRFA, each modality will {create} a memory block $\phi$ that stores unimodal semantic features. After recursive strengthening of the memory blocks, a feature sequence $X_m^\phi$  of size $\left(l_{m}^{\phi}\times d_{m}^{\phi}\times b\right)$ is obtained. Here, $l_m^\phi$ and $d_m^\phi$ sub-tables represent each memory entry's length and original feature dimension, and $b$ is the batch size.
	
{Then,} MRFA uses the semantic features $X_m^h$ extracted from different unimodal feature extractors to generate a memory block by the subsequent recurrent alignment augmentation network. As shown in Eq. (6), $X_m^\phi$ uses the features $X_{m^\prime}^h$ and $X_{m^{\prime\prime}}^h$ of the other two modalities for alignment refinement.
\begin{equation}
	X_m^\phi= \text{{MRFA}}(X_m^h,X_{m'}^h,X_{m''}^h)
\end{equation}
	
In subsequent model runs, MRFA reads the content from the memory repository, continuously refines $X_m^\phi$ using $X_m^h$, and aligns the unimodal latent semantic information with the other two modalities. These refined unimodal semantic features obtained above will be stored in the memory block to replace the old content in the block.
	
Moreover, MRFA uses an attention mechanism for each modality to extract significant semantic features $X_m^R{\in\mathbb{R}}^{b\times F_m^R}$. {Here,} $F_m^R$ represents the embedding size of the enhanced unimodal feature. Intra-modality memory attention performs refined feature extraction on the semantic features $X_m^\phi$ of each entry $\tau$ in the three modal memory blocks. The calculation formula of the attention weight of each memory entry $\tau$ in the memory blocks is as follows:
\begin{equation}
	\mu_{(m,\tau)}=W_m^{R^T}X_{(m,\tau)}^\phi
\end{equation}
\begin{equation}
	\omega_{(m,\tau)}=\frac{\exp(\mu(m,\tau))}{\sum_{\tau}^{l_m}\ \exp(\mu(m,\tau))}
\end{equation}
	
Then, MRFA utilizes $\omega_{(m,\tau)}$ to fuse memory block features to extract the salient unimodal representations in each modality. $W_m^{R^T}$ represents a learnable dynamic parameter. {$l_m$ indicates the number of entries.}
\begin{equation}
	X_m^R=\sum_{\tau=1}^{{l_m}}\ \omega_{(m,\tau)}X_{(m,\tau)}^\phi
\end{equation}

In MRFA, we use {the lightweight model 1D-CNN to calculate the attention weight and set the filter size to 1 to speed up the attention calculation.}

\subsubsection{Cross-modal Multi-head Attention}
First, we apply linear projections to the query $Q\in\mathbb{R}^{T_Q\times{d}_Q}$, key $K\in\mathbb{R}^{T_K\times{d}_K}$, and value $V\in\mathbb{R}^{T_V\times{d}_V}$, $n$ times using distinct linear transformations. Here, $n$ represents the number of heads. $T_Q$, $T_K$, and $T_V$ denote the sequence lengths of the query, key, and value, respectively, representing the number of elements in each sequence. Similarly, $d_Q$, $d_K$, and $d_V$ are used to indicate the feature dimensions of the query, key, and value, respectively, representing the number of features or dimensions in each element of the sequences.

{After the basic description, to precisely illustrate how the cross-modal multi-head attention mechanism is applied to data of different modalities, we will use one modality as an example for illustration. Let $m$ represent one of the modalities, such as text, and $m'$ represent one of the other two modalities, such as visual or audio. In this case, we first process the text modality ($m$) data, applying $n$ different linear transformations to generate the query $\hat Q$. We then apply $n$ different linear transformations to the other modality ($m'$) data to generate $\hat K$ and $\hat V$. This way, we can capture cross-modal interaction information between each modality and other modalities. The specific calculation is as follows:}

\begin{equation}
	\hat Q = \text{Concat}(QX_m^RW_1^Q, \ldots ,QX_m^RW_i^Q, \ldots ,QX_m^RW_n^Q)
\end{equation}
\begin{equation}
	\hat K = \text{Concat}(K{X_{m'}^R}W_1^K, \dots, K{X_{m'}^R}W_i^K, \dots, K{X_{m'}^R}W_n^K)
\end{equation}
\begin{equation}
	\hat V = \text{Concat}(V{X_{m'}^R}W_1^V, \dots, V{X_{m'}^R}W_i^V, \dots, V{X_{m'}^R}W_n^V)
\end{equation}
where $W_i^Q\in\mathbb{R}^{d_Q\times d_m}$, $W_i^K\in\mathbb{R}^{d_K\times d_m}$ and $W_i^V\in\mathbb{R}^{d_V\times d_m}$ {are learnable parameters} in the fully connected layer, and $d_m$ is the output dimension.
	
Then, we use the dot-product attention to compute queries ${QW_i^Q}\in\mathbb{R}^{T_Q\times d_m}$, keys ${KW_i^K}\in\mathbb{R}^{T_K\times d_m}$, and values ${VW_i^V}\in\mathbb{R}^{T_V\times d_m}$ on each projection, and get attention scores for feature vectors composed of different relations. The formula is defined as follows:
\begin{equation}
	hea{d_i} = \text{{softmax}}\left( {(\hat QW_i^Q){{(\hat KW_i^K)}^T}} \right)(\hat VW_i^V)
\end{equation}
	
Next, the outputs of the attention function $head_i,i\in{[1,n]}$ are concatenated to form the final value $X_m^{head}$, which is calculated as follows:
\begin{equation}
	X_m^{head} = \text{{Concat}}(head_1, head_2, \dots,head_n)
\end{equation}
	
Here, $head_i\in\mathbb{R}^{T_Q\times d_m}$ and $X^{head}\in\mathbb{R}^{T_Q\times nd_m}$. Finally, we employ a GRU model to capture the correlation between feature alignments at each iteration. The formula is defined as follows:
\begin{equation}
	{X_m} = \text{{GRU}} \left( {{X_m},X_m^{head}} \right)
\end{equation}
	
\subsection{Multimodal Fusion with masked GCN}
In this section, we introduce the embedding of speaker information, the process of graph construction, and the masked graph mechanism.
	
\subsubsection{Speaker Embedding}Some existing GCN models do not consider embedding learning of speaker information when constructing graphs, resulting in the inability to use speaker information to model potential connections within or between speakers. To solve the above problems, this paper embeds speaker information into GCN. Assuming that there are $N$ dialogue characters in the data set, then the size of our speaker embedding is also $N$. We show the speaker information embedding process in Fig. \ref{fig_2}. The original speaker information can be represented by the vector $S_i$, and $ X^S$ represents the speaker's embedding. The calculation process is as follows:
\begin{equation}
	X^S=W_sS_i+X_m^{head}
\end{equation}
	
Here, $W_s$ represents a learnable weight matrix.
After the above process, we embed the speaker information in GCN modeling with additional information about the speaker.
	
\subsubsection{Graph Construction}
The graph construction process consists of node representation, {edge connection}, and edge weight initialization. The following will introduce each in detail.
	
\paragraph{Node Representation}
We form a graph of text, visual, and audio data, and it is expressed as $\mathcal{G}_{m}=\left\{\mathcal{V}_{m},\mathcal{A}_{m}, \mathcal{E}_m\right\}$. Where $\mathcal{V}_m$ represents the node sets, {$\mathcal{A}_m\in\mathbb{R}^{|\mathcal{V}_m|\times|\mathcal{V}_m|}$} is the adjacency matrix, {and} $\mathcal{E}_m$ {represents} the edge sets. Any node $v_m^i\in \mathcal{V}_m$ in the graph contains a sentence in modalities. At this time, each sentence node $v_m^i$ in the fusion graph contains the semantic features $X^S\in R^{d_{m}}$ from the three modal fusions.
	
\paragraph{Edge Connection}In the same dialogue, we assume that there are explicit or latent connections between arbitrary sentences. Therefore, in the graph constructed in this study, any two nodes in the same modality in the same dialogue are connected. Furthermore, each node is also connected to nodes in the same conversation in different modalities due to the complementarity of the same discussion among other modalities.
	
\paragraph{Edge Weight Initialization}The graph designed in this study has two different types of edges. (1) the two nodes connected by the edge come from the exact modal; (2) the two nodes connected by the edge come from two different {modes}. In order to capture the similarity of node representations, we employ degree similarity to determine the edge weight. The significance of an edge connecting two nodes is directly proportional to their similarity. This implies that nodes with higher similarity exhibit more crucial information interaction between them.
	
To handle different types of edges, we use different edge weighting strategies. For edges from the same modality, our approach is computed as follows:
\begin{equation}
	\mathcal{A}_{ij}=\mathbf{1}-\frac{\text{arccos}\left(\text{{sim}}(n_in_j)\right)}{\pi}
\end{equation}
where $n_i$ and $n_j$ represent the feature representations of the $n$-th and $j$-th nodes in the graph. For edges from the different modalities, our approach is computed as follows:
\begin{equation}
	\mathcal{A}_{ij}=\aleph\left(1-\frac{\arccos\left(\text{{sim}}(n_i,n_j)\right)}{\pi}\right)
\end{equation}
where $\aleph$ is a {hyperparameter}.
	
\begin{figure}[htbp]
	\centering
	\includegraphics[width=1.0\linewidth,scale=1.00]{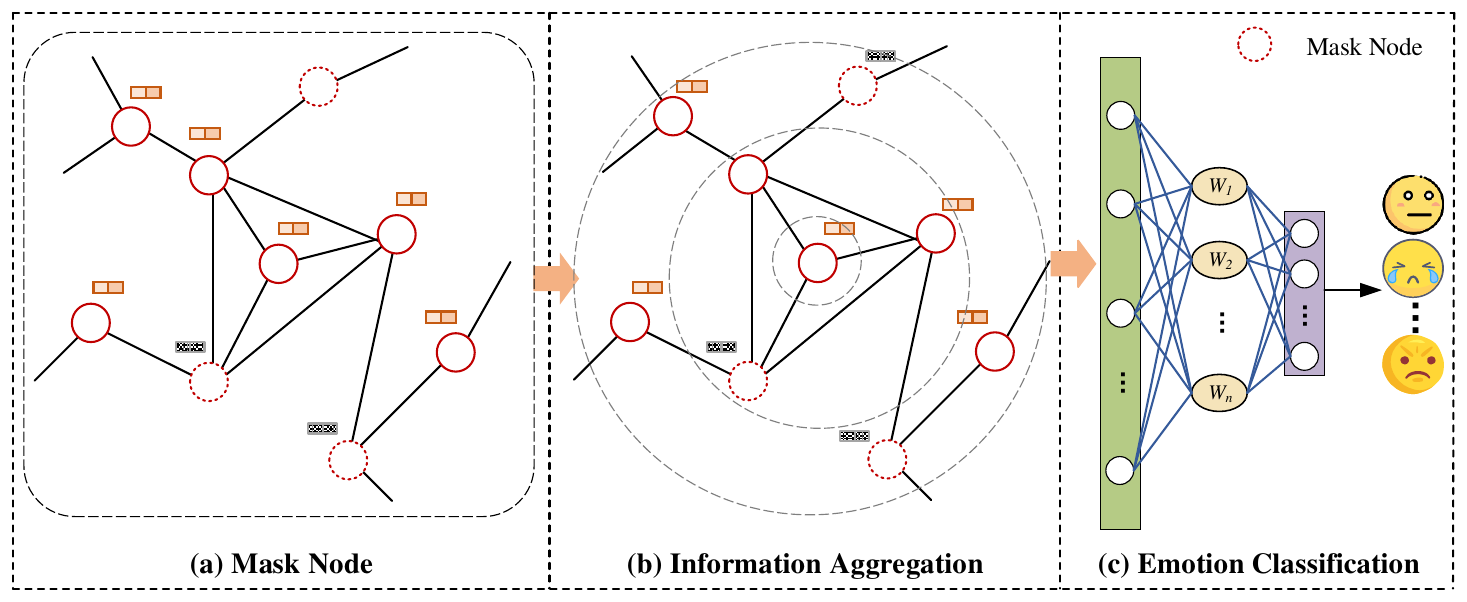}
	\caption{Randomly mask the nodes on the graph, and use GCN for information aggregation to achieve the final fusion of multimodal features and emotion classification.}
	\label{fig_4}
\end{figure}
	
\subsubsection{Graph Learning and Mask Mechanism}
Through the analysis and research of GCNs used in the past for multimodal emotion recognition, we found that the fully connected graphs constructed by most previous models have more or less introduced redundant or noisy information in the fusion process. The node structure in graph information is overemphasized, and they usually build more edges on GCNs to exploit the topological closeness between adjacent nodes. Based on this situation, we utilize a random masked graph neural network, as shown in Fig. \ref{fig_4}, by randomly masking the adjacency matrix to remove excessive noise in the model and improve the robustness of GCN. In addition, our masked GCN can also reduce the number of network parameters of the model on a large scale so that the model avoids the problem of overfitting.
	
We achieve the above effect by exploiting a random mask on the adjacency matrix. Mathematically, a subset $\widehat{\mathcal{V}}\subset{\mathcal{V}}$ of nodes is sampled, a subset $\widehat{{\mathcal{E}}}\subset{{\mathcal{E}}}$ of nodes is sampled and a mask mark $[M]$ is defined.
\begin{equation}
	x_i=\left\{\begin{matrix}x_{[M]}\quad &,v_i\in\stackrel{\sim}{\mathcal{V}}\\x_i&,v_i\notin\stackrel{\sim}{\mathcal{V}}\end{matrix}\right\}.
\end{equation}
\begin{equation}
	e_i=\left\{\begin{matrix}e_{[M]}\quad &,e_i\in\stackrel{\sim}{E} \\e_i&,e_i\notin\stackrel{\sim}{E}\end{matrix}\right\}.
\end{equation}
	
Here $x_i$ and $e_i$, respectively, represent the set of nodes and edges after the mask.

We construct a masked GCN, which is used to exploit the semantic complementarity between different modalities further to encode context dependencies. Specifically, given an undirected graph $\mathcal{G}_{m}=\{\mathcal{V}_{m},\mathcal{A}_{m}, \mathcal{E}_m\}$,  $\tilde{\mathcal{P}}$ be the renormalized graph laplacian matrix of {$\mathcal{G}_{m}$}:
\begin{equation}
	\begin{aligned}
			\overset{\sim}{\mathcal{P}} &= \overset{\sim}{\mathcal{D}}^{-1/2}\overset{\sim}{\mathcal{A}}\mathcal{D}^{-1/2} \\
			&= (\mathcal{D}+\mathcal{I})^{-1/2}(\mathcal{A}_{[M]}+\mathcal{I})(\mathcal{D}+\mathcal{I})^{-1/2}
	\end{aligned}
\end{equation}
	
Here, $\tilde{\mathcal{P}}$ represents a learnable weight matrix, $\mathcal{A}$ represents the neighbor weight matrix, $\mathcal{A}_{[M]}$ represents the mask matrix, $\mathcal{D}$ represents the diagonal matrix, and $\mathcal{I}$ represents the identity matrix.

\subsection{Emotion Classifier}
Using the masked GCN, we fuse the semantic information from multiple modalities to obtain the semantic representation of various modalities.
	
Then we input the feature $\tilde{\mathcal{P}}$ that combines multiple modal semantic information into an MLP with a fully connected layer, then use the RELU activation function for nonlinear activation, and normalize the feature information $P_i$ of the hidden layer through the Softmax function:
\begin{align}
	l_i&=\text{{RELU}}(W_l\tilde{\mathcal{P}}+b_l)\\
	P_i&=\text{{Softmax}}\left(W_{smax}l_i+b_{smax}\right)
\end{align}
	
Here, $W_l$ and $W_{smax}$ represent a learnable weight matrix. Finally, we use the argmax function to match $P_i$ with the emotional label $\hat{y_i}$ of the utterance and use the formula to express the process of predicting the emotional label $\hat{y_i}$ of the utterance as follows:
\begin{equation}
	\overset{\wedge}{y_i}=\text{{argmax}}\left(P_i[k]\right)
\end{equation}
	
The entire inference process of the MGLRA pseudocode is contained in Algorithm \ref{alg1}.

\begin{algorithm}[H]
	\caption{MGLRA}\label{alg:alg5}
	\begin{algorithmic}[1]		
	\item[] \textbf{Input}:
	$X_{t}^{r},X_{a}^{r},X_{v}^{r}$ $\leftarrow$ RoBERTa,OpenSMILE,3D-CNN
	\item[]\hspace{1cm} $T^F$: total MRFA iterations
\item[] \textbf{Output}: $\overset{\wedge}{y_i}$: emotion label
		
        \STATE {\textbf{{for}}} {$epoch \leftarrow 1,$ $ 2, ... ,$ $n$} {\textbf{{do}}}
		\STATE \hspace{0.5cm}$X_t^c,X_a^c,X_v^c \leftarrow {\text{LSTM}}({X_t^r,X_a^r,X_v^r})$	
		\STATE \hspace{0.5cm}$\hfill$$\triangleright$ (Eq. 2)
		\STATE \hspace{0.5cm}$X_{m}^{h} \leftarrow$ Graph Attention Filtering Mechanism($X_{m}^{c}$)
		\STATE \hspace{0.5cm}$\hfill$$\triangleright$ (Eqs. 3 \& 4 \& 5)
		
		\STATE \hspace{0.5cm}{\textbf{for}} {$k \leftarrow 1,$ {2, ...,} $T^F$} {\textbf{do}}
		\STATE \hspace{1.0cm}{\textbf{Repeat steps 6-25 for each modality }}
		\STATE {\hspace{1.0cm}{\textbf{if}} $k = 1$ {\textbf{then}}}
		\STATE {\hspace{1.5cm}$X_m^\phi\leftarrow$Memory Module $(X_m^h,X_{m'}^h,X_{m''}^h)$}
		\STATE {$\hfill$$\triangleright$ Initialize memory block with $X_m^h$}
		\STATE {\hspace{1.0cm}{\textbf{else}}}
		\STATE {\hspace{1.5cm}$X_m^\phi\leftarrow$Memory Module $(X_m)$}
		\STATE {$\hfill$$\triangleright$ Use $X_m$ for subsequent rounds}
	    \STATE {\hspace{1.0cm}{\textbf{end if}}}
		\STATE \hspace{1.0cm}Calculate attention $\omega_{(m,\tau)}$ for each entry of $X_m^\phi$
		\STATE $\hfill$$\triangleright$  (Eqs. 7 \& 8)
		\STATE \hspace{1.0cm}$X_m^R=\sum_{\tau=1}^{l_m}\ \omega_{(m,\tau)}X_{(m,\tau)}^\phi$
		\STATE $\hfill$$\triangleright$  Unimodal feature (Eq. 9)
		
		\STATE \hspace{1.0cm}Project query  ($\hat{Q}$),  key  ($\hat{K}$) and value ($\hat{V}$)
		\STATE $\hfill$$\triangleright$   (Eqs. 10 \& 11 \& 12)
		\STATE \hspace{1.0cm}$X_m^{head}\leftarrow {\text{Concat}} (head_1, head_2, \dots,head_n)$
		\STATE $\hfill$$\triangleright$ (Eqs. 13 \& 14)	
	
		\STATE \hspace{1.0cm}${X_m} \leftarrow {\text{GRU}} ( {{X_m},X_m^{head}} )$
		\STATE $\hfill$$\triangleright$ (Eq. 15)
		
		\STATE \hspace{0.5cm}{\textbf{end for}}	
		
		\STATE \hspace{0.5cm}$X^S=W_sS_i+X_m^{head}$
		\STATE \hspace{0.5cm}$\hfill$$\triangleright$   Speaker Embedding (Eqs. 16)
		\STATE \hspace{0.5cm}$\tilde{\mathcal{P}} \leftarrow $masked GCN$(X^S)$
		\STATE \hspace{0.5cm}$\hfill$$\triangleright$ (Eqs. 17 \& 18 \& 19 \& 20 \& 21)
		\STATE \hspace{0.5cm}$\overset{\wedge}{y_i} \leftarrow {\text{RELU}} (\tilde{\mathcal{P}})+{\text{Softmax}}(l_i) $	\STATE$\hfill$$\triangleright$ (Eqs. 22 \& 23)
        \STATE {\textbf{{end for}}}
		\STATE {\textbf{return}} $\overset{\wedge}{y_i}$
	\end{algorithmic}
	\label{alg1}
\end{algorithm}

\section{Experiments}
\label{sec:experiments}
This section introduces two commonly used datasets for multimodal emotion recognition and the evaluation indicators of related experiments. We show our setup and experimental procedure on these two datasets and discuss and analyze our comparison methods and results. At the same time, we also checked the results of the ablation experiments. We use visualization experiments to study the distribution of semantic features to verify the effectiveness of our proposed method.
	
\subsection{Benchmark Datasets }
Based on the latest research results in the MERC field, we selected two benchmark datasets of different scales, IEMOCAP\cite{31} and MELD\cite{36}, to conduct experiments to verify the innovation and performance of our proposed algorithm model. The detailed visualization statistics of the two datasets are shown in Table \ref{tab1}.
	
	\begin{figure}[htbp]
		\begin{minipage}[t]{0.49\linewidth}   
			\centering
			\includegraphics[width=\linewidth,scale=1.00]{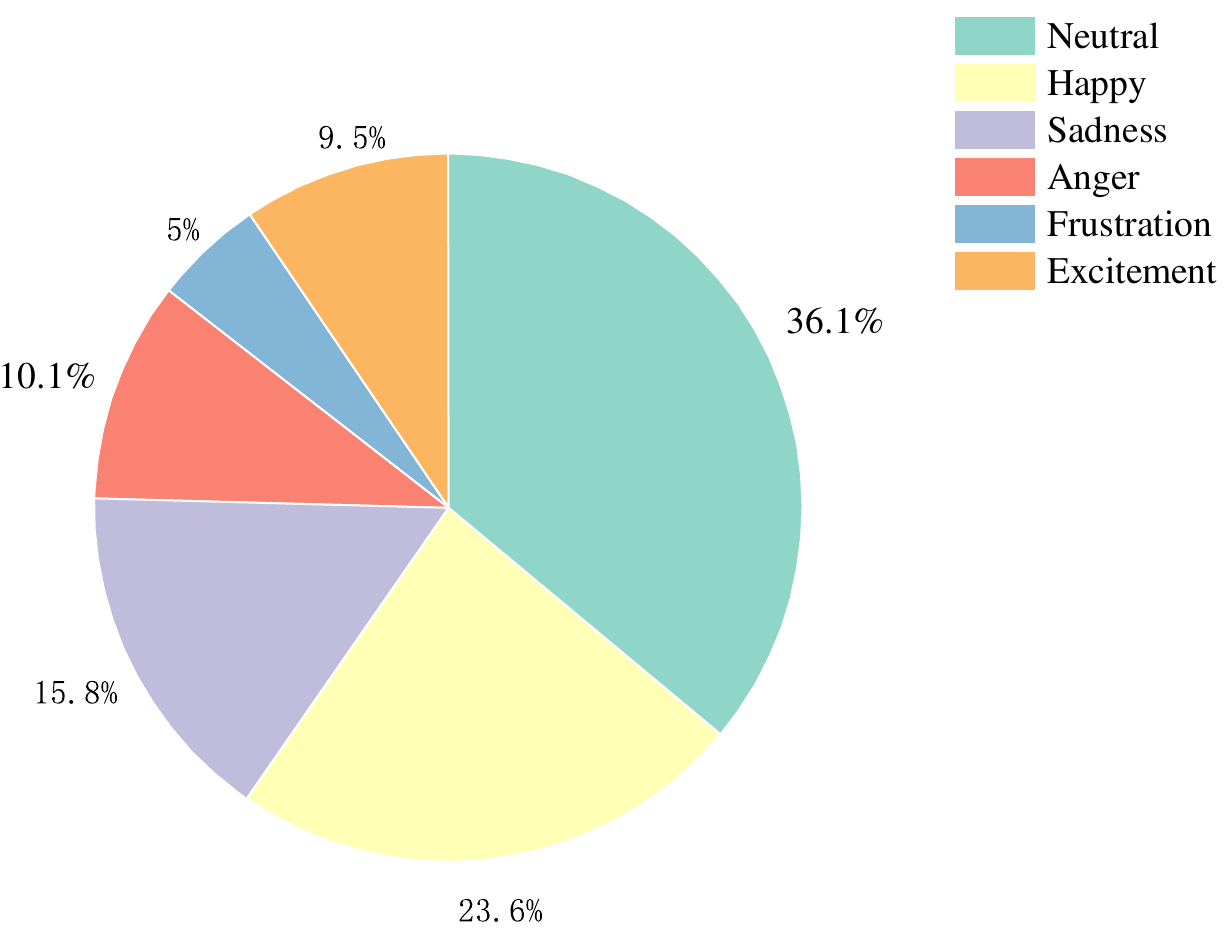}
			\scriptsize
			\\ (a) IEMCOAP
		\end{minipage}
		\begin{minipage}[t]{0.49\linewidth}
			\centering
			\includegraphics[width=\linewidth,scale=1.00]{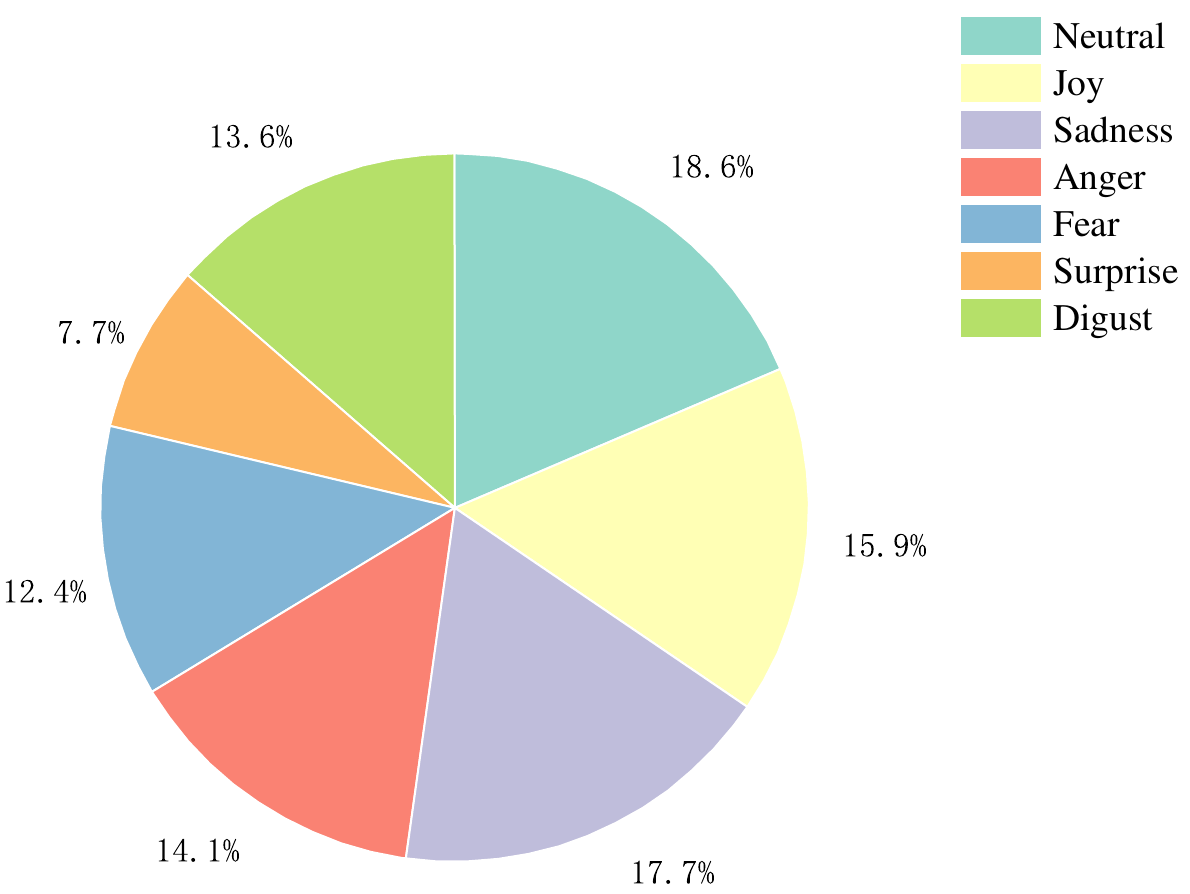}
			\scriptsize
			\\ (b) MELD
			\label{fig:image2}
		\end{minipage}
		
		\caption{Emotion label distribution on IEMOCAP and MELD datasets. Compared with MELD's emotional label distribution, IEMOCAP has a severe data imbalance problem, indicating that it is more difficult to identify during the experiment.}
		\label{fig_5}
	\end{figure}

\textbf{IEMOCAP} is a multimodal dataset of binary interactions between actors consisting of five males and five females developed at the University of Southern California. The dialogue process between them was recorded using a video camera and a motion capture device. The dialogue recording contained five groups, each group had a boy and a girl, and each group of interactive sessions lasted 5 to 10 minutes. IEMOCAP provides the actors' video, transcription, and audio in detail. In addition, the dataset also includes annotations of various other features, such as facial expressions and head and body movements. This consists of the script content and some improvisational performances to restore the multiple situations in which emotions occur more realistically. IEMOCAP divides emotion tags into six categories: happy, anger, frustration, sadness,  excitement, and neutral. The distribution of emotion labels is shown in Fig. \ref{fig_5}(a). A total of {7433} samples were used for experimental verification.
	
\begin{table*}[!t]
\renewcommand\arraystretch{1.5}
\caption{Detailed visualization statistics for iemocap and meld datasets.}
\setlength\tabcolsep{4mm}{
	\begin{tabular}{c|clc|clc|c|c|c}
		\hline
		\multirow{2}{*}{Dataset} & \multicolumn{3}{c|}{Utterance Count}          & \multicolumn{3}{c|}{Dialogue Count}           & \multirow{2}{*}{Number Of Speakers} & \multirow{2}{*}{Total Duration(Hours)} & \multirow{2}{*}{Evaluation Metrics} \\ \cline{2-7}
		& \multicolumn{2}{c|}{train+val} & test         & \multicolumn{2}{c|}{train+val} & test         &                                     &                                        &                                     \\ \hline
		IEMOCAP                  & \multicolumn{2}{c|}{5810}      & 1623         & \multicolumn{2}{c|}{120}       & 31           & 10                                  & 12.5                                   & Accuracy/F1                         \\ \hline
		MELD                     & \multicolumn{2}{c|}{11098}     & 2610         & \multicolumn{2}{c|}{1153}      & 280          & 5                                   & 73.3                                   & Accuracy/F1                         \\ \hline
\end{tabular}}
\label{tab1}
\end{table*}

\textbf{MELD} is a collection of audiovisual clips designed for research in emotion recognition. The dataset contains 1,535 pins from the classic TV series \textit{Friends}, each roughly 3-5 seconds long. The clips were labeled with seven basic emotions: anger, disgust, fear, happiness, sadness, and surprise, as well as a neutral category. MELD provides data in visual, audio, and text form for experiments. Unlike IEMOCAP, in addition to emotion labels, the dataset includes demographic information about the actors in the clips, such as their gender, age, and race. Another difference from IEMOCAP is that MELD divides the dataset into three parts for users: the development set, training set, and test set, thus providing a unified test method when analyzing the model. The distribution of emotion labels is shown in Fig. \ref{fig_5}(b).

{Furthermore, {following previous work \cite{9}, \cite{17}, \cite{li2023graphcfc}, we adopt a standard train/test split method to separate the training, validation, and test sets. Specifically, we use the first four sessions of the dataset as the training and validation sets and the last session as the test set. This approach ensures that the model's performance is evaluated on different sets, thereby maximizing the diversity of training and testing scenarios and ensuring that the specific characteristics of any single session do not bias the model's performance.}}

\subsection{Implementation Details}
Our experiments were conducted on a GeForce RTX 3090 GPU server with 24 GB of memory using an Intel Core 12900K CPU. The experimental code was developed based on the PyTorch 1.8.1 deep learning framework and implemented using Python 3.7. {To obtain the best performance of the MGLRA model, we set the number of heads of the multi-head attention mechanism to 10 and the random masking rate of the masked GCN to 0.7. The batch size of the model was set to 32, and the model was trained for 70 epochs, each taking approximately 55 seconds. We use the Adam optimizer and set the learning rate to 0.0001 and weight decay to 0.00005.} Due to the varying data sizes of the IEMOCAP and MELD datasets, we train the models for 3000 steps and 2500 steps, respectively. During training, the models are evaluated or tested every 20 steps, and the best-performing model is saved.

\subsection{Baselines and Evaluation Metrics}

To verify the effectiveness of our model on MGLRA, the paper compared the following baseline models with our model: text-CNN \cite{3}, MFN \cite{32}, bc-LSTM \cite{33}, CMN \cite{34}, DialogueRNN \cite{16}, DialogueGCN \cite{7}, ICON \cite{35}, A-DMN \cite{xing2020adapted}, CTnet \cite{9}, LR-GCN \cite{17}, GraphCFC \cite{li2023graphcfc}. {It should be noted that since the DialogueGCN and LR-GCN methods cannot directly fuse multi-modal information, we extended them through a linear fusion layer. The extended methods are represented by DialogueGCN* and LR-GCN*, respectively.}
	
\begin{table*}[!t]
\renewcommand\arraystretch{1.5}
\setlength{\tabcolsep}{14.8pt}
\caption{{Our proposed model is evaluated on the IEMOCAP dataset, and its performance is compared with various baseline and state-of-the-art models. The IEMOCAP dataset consists of six labels, and we measure the performance using Acc.($\%$) and F1-score, which represents the accuracy score and weighted average score, respectively. The models that excel in vertical contrast are highlighted in bold.}}

\begin{tabular}{l|ccccccc}
	\hline
	   \multirow{3}{*}{Methods} & \multicolumn{7}{c}{IEMOCAP}  \\ \cline{2-8}
	   & Happy      & Sadness        & Neutral    & Angry      & Excitement    & Frustration & Average(w) \\ \cline{2-8}
	   & Acc.  F1   & Acc.  F1   & Acc.  F1   & Acc.  F1   & Acc.  F1   & Acc.  F1   & Acc.  F1   \\ \hline
	   bc-LSTM                  & 29.2  34.3 & 57.2  60.9 & 54.2  51.9 & 57.1  56.8 & 51.2  58.0 & 67.2  58.0 & 55.3  55.1 \\
	   CMN                      & 25.2  30.4 & 56.1  62.3 & 52.9  52.4 & 61.6  59.9 & 55.4  60.1 & \textbf{71.2}  60.5 & 56.6  56.0 \\
	   ICON                     & 22.3  30.1 & 58.9  64.7 & 62.8  57.3 & 64.8  63.1 & 59.0  63.2 & 67.3  60.9 & 59.2  58.6 \\
	   MFN                      & 23.9  34.2 & 65.2  70.4 & 55.4  51.9 & 71.9  66.9 & 64.2  61.9 & 68.1  62.4 & 60.2  60.1 \\
	   DialogueRNN              & 25.5  33.2 & 75.2  78.7 & 58.4  59.3 & 64.8  65.3 & 80.1  71.7 & 61.2  59.0 & 63.5  62.6 \\
	   A-DMN                   & 43.2  50.7 & 69.5  76.9 & 63.2  63.1 & 63.4  56.7 & \textbf{88.2}  77.8 & 53.4  55.8 & 64.8  64.3 \\
	   DialogueGCN              & 40.4  42.6 & {89.2  84.5} & 62.0  63.4 & 67.4  64.2 & 65.5  63.1 & 64.0  66.8 & 65.1  64.2 \\
       {DialogueGCN*}                 & {40.1  43.1} & \textbf{{89.6  85.0}} & {61.5  62.9} & {67.0  63.9} & {66.0  63.6} & {63.6  66.2} & {65.5  63.9} \\
	   CTnet                   & 48.2  51.2 & 78.1  79.9 & 69.1  65.7 & \textbf{72.8} 67.3 & 85.4  \textbf{78.4} & 52.3  58.9 & 68.1  67.6 \\
	   LR-GCN                  & 54.3  55.6 & 81.7  79.0 & 59.3  63.9 & 69.5  69.1 & 76.2  74.1 & 68.1  {68.7} & 68.4  68.2 \\
	   {LR-GCN*}                   & {55.8 56.9} & {82.4 79.5} & {60.3 64.6} & {70.2 69.8} & {76.8 74.7} & {68.9} \textbf{{69.3}} & {68.8 68.6} \\	
	   GraphCFC                  & 43.2  54.2 & 85.0  84.3 & 64.6  62.1 & 71.4 \textbf{70.2} & 78.9  74.1 & 63.8  62.1 & 69.1  68.5 \\
	   MGLRA                   & \textbf{62.9  63.5} & 81.1  81.5 & \textbf{70.9 71.5}   & 60.2  61.1 & 74.4  76.3 & 69.2  67.8 & \textbf{71.3  70.1} \\ \hline
\end{tabular}
\label{tab2}
\end{table*}
	
To compare with other studies, we uniformly use the accuracy and F1-score to evaluate the performance of our model and use the weighted average method to reduce the error.

\section{Results and Discussion}
\label{sec:resultsdiscussion}
\subsection{Comparison with State-of-the-Art and Baseline Methods}
We evaluate the performance of the proposed model and compare it with baseline methods and state-of-the-art methods. Table \ref{tab2} represents the performance of all models on the IEMOCAP dataset, while Table \ref{tab3} represents the performance of all models on the MELD dataset.
	
\begin{table*}[!t]
	\centering
	\renewcommand\arraystretch{1.5}
	\caption{{Our proposed model is evaluated by comparing its performance with various baseline and state-of-the-art models on the MELD dataset. The MELD dataset consists of seven labels. We assess the performance using Acc. ($\%$) and F1-score, which indicate the accuracy score and weighted average score, respectively. Those that perform best in vertical contrast are highlighted in bold.}}
	\setlength{\tabcolsep}{12.2pt}{
		\begin{tabular}{l|*{8}{c}}
			\hline
			\multirow{3}{*}{Methods} & \multicolumn{8}{c}{MELD} \\ \cline{2-9}
			& Neutral & Surprise & Fear & Sadness & Joy & Disgust & Anger & Average(w) \\
			& Acc. F1 & Acc. F1 & Acc. F1 & Acc. F1 & Acc. F1 & Acc. F1 & Acc. F1 & Acc. F1\\ \hline
			text-CNN & 74.8 73.2 & 45.4 44.8 & 3.8 3.2& 21.4 22.3& 49.3 48.9 & 8.9 8.7 & 35.3 35.1& 55.3 54.9 \\
			MFN & 76.4 76.2 & 40.9 39.5 & 0.0 0.0 & 14.1 13.7 & 46.7 45.8 & 0.0 0.0 & 40.9 39.6& 55.1 54.2 \\
			bc-LSTM & 77.0 76.2& 25.1 24.3 & 9.2 8.7  & 24.4 23.5& 54.6 54.0& 4.4 4.2& 43.7 42.8& 59.3 58.3\\
			CMN & 74.7 73.4& 46.8 45.8 & 0.0 0.0 & 23.4 22.3 & 45.0 44.5& 0.0 0.0& 44.8 43.7& 55.7 54.2\\
			ICON & 73.8 73.5 & 50.0 48.9& 0.0 0.0 & 23.4 22.3& 50.6 49.8 & 0.0 0.0& 45.0 43.7& 56.5 55.2 \\
			DialogueRNN & 77.4 78.2& 52.8 51.9& \textbf{11.9} \textbf{10.8}& 34.5 33.8& 54.4 52.9& 7.8 6.9& 43.7 42.5& 60.1 59.4\\
			A-DMN & 78.8 77.6 & 55.4 54.9 & 8.7 6.8& 24.7 23.9& 24.6 22.7 & 3.5 3.2 & 41.3 40.9& 55.5 55.4\\
			GraphCFC & 76.6 75.8 & 49.5 48.5 & 0.0 0.0 & 26.7 25.9& 52.2 52.1 & 0.0 0.0& 47.5 47.2 & 61.5 61.3\\
			LR-GCN & \textbf{80.1} 80.5 & 57.2 56.8 & 0.0 0.0& {36.9} {35.8}& 65.8 64.7& \textbf{11.1} \textbf{10.3} & {54.3} {53.2}& 65.7 {65.2} \\
			{LR-GCN*} & {79.8 80.3} & {57.5 57.0} & {0.0 0.0} & \textbf{{37.2}} \textbf{{36.3}} & {66.1 65.2} & {{10.9} {10.7}} & \textbf{{54.7}} \textbf{{53.9}} & {65.9} \textbf{{65.3}}\\
			MGLRA & 78.2 \textbf{80.8} & \textbf{59.8} \textbf{59.5} & 0.0 0.0& 30.8 27.8 & \textbf{68.5} \textbf{66.5}& 0.0 0.0& 43.6 48.4& \textbf{66.4} {64.9} \\ \hline
	\end{tabular}}
\label{tab3}
\end{table*}
	
\textbf{IEMOCAP:} Observing Table \ref{tab2}, GraphCFC has the best performance among all baseline models. GraphCFC extracts all the different edge types from the constructed graph for further encoding, enabling GCN to model the interaction between contexts in semantic information transfer accurately. We align semantic information using a recurrent alignment network and a more lightweight GCN to incorporate multimodal semantic details in our work, resulting in better performance than GraphCFC. {Among all the compared models, LR-GCN* achieves excellent performance second only to our model and GraphCFC, with 68.8\% accuracy and 68.6\% F1-score, respectively, effectively utilizing the latent contextual semantic session information.} LR-GCN introduced a multi-head attention mechanism to find potential connections between utterances. It then encoded the enhanced multimodal semantic information into a fully connected graph, and this approach also inspired our work. {However, both GraphCFC and LR-GCN focus on the fusion of later models, ignoring the noise within and between multimodal features in the early alignment process, which limits their performance. Compared with LR-GCN* and GraphCFC, our model improves weighted accuracy and F1-score by 2.5\%, 2.2\%, 1.5\%, and 1.6\%, respectively.} Our model achieves the maximum performance across six metrics on the IEMOCAP dataset, and has a more balanced performance on each emotion label than other models.
	
\textbf{MELD:} As can be seen from Table \ref{tab3}, {the overall performance of our MGLRA model is relatively close to that of LR-GCN*. However, it surpasses the *LR-GCN method in two emotion categories and overall weighted average performance. Specifically, across all methods listed, MGLRA achieved an accuracy of 66.4\%, an F1 score of 64.9\%, an accuracy of 59.8\% for the surprise category, and an accuracy of 68.5\% for the joy category. Using graph attention filtering mechanism and recurrent alignment architecture to align the three modes before fusion can achieve better results. However, *LR-GCN performs well in the Digust emotion category and achieves state-of-the-art performance .*LR-GCN uses multi-head attention to dynamically explore potential relationships between sentences and introduces densely connected graphs to further capture graph structure information. This is more effective for the MELD data set with imbalanced data and can better capture the variety of small sample emotions.}
	
{Experimental results show that our proposed MGLRA model has absolute advantages over the IEMOCAP data set. It is on par with or slightly ahead of the LR-GCN model on the MELD data set.} In general, MGLRA can more effectively eliminate noise and enhance semantic features during the modal alignment process. {It can also} effectively utilize this information to improve the overall performance of the model. {This improvement is mainly attributed to two reasons: (1) MGLRA aligns the three modalities using the graph attention filtering mechanism and iterative augmentation architecture, thereby capturing more emotional information for fusion. (2) During the fusion process, masked GCN randomly discards some nodes while incorporating speaker information. These advantages improve the experimental performance of our model on the MELD dataset.}

	\begin{figure}[htbp]   
	\centering
	\includegraphics[width=0.49\textwidth,scale=1.00]{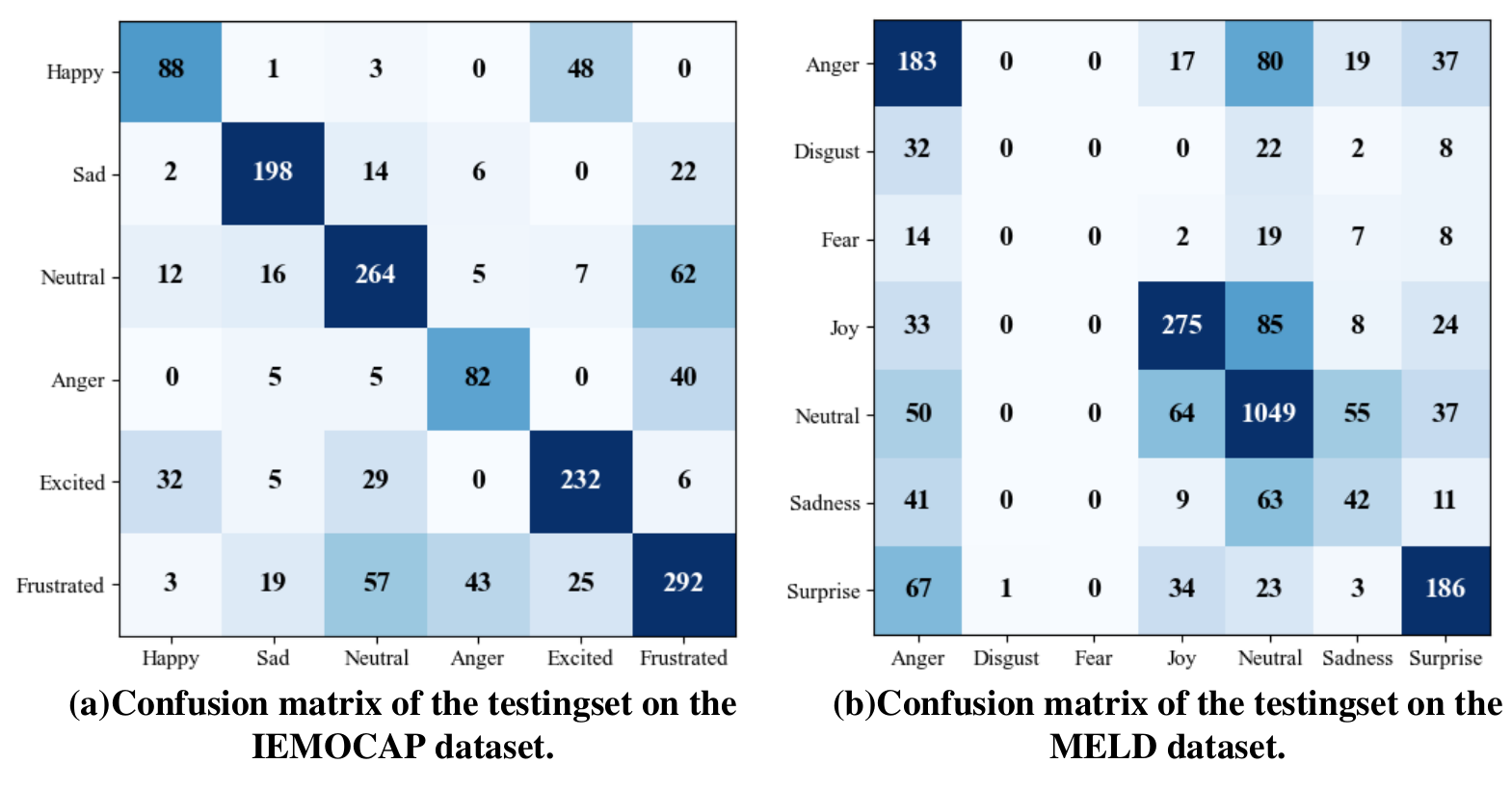}
	\caption{The confusion matrices present the true and predicted labels of the testing set for both the IEMOCAP dataset and the MELD dataset. The columns represent the true labels, while the rows represent the predicted labels.}
	\label{fig_6}
\end{figure}
	
To analyze the performance of MGLRA more comprehensively, we show the confusion matrices on the IEMCOAP and MELD datasets in Fig. \ref{fig_6}. By analyzing the confusion matrix of {IEMOCAP, it} can be seen that people's expressions of excitement and happiness are very similar. The reason is that the activation domain and valence domain of excitement and {happiness} are relatively close, which leads to the model making a confusing judgment. It is easy to misclassify the two emotions. By analyzing the confusion matrix of MELD, it can be seen that because the data set does not provide adequate training data, our model is almost unable to recognize the emotions of disgust and fear. By analyzing the distribution of different emotion categories in the MELD dataset, we know 361 sample data of disgust emotion categories in MELD. Still, there are only 68 samples in the test set, and what is more serious is that there are only 50 samples in the test set for the fear category. The neutral class has up to 1256 samples in the test set. Due to the severe sample distribution imbalance problem in the MELD dataset, there are considerable differences in the model's performance in different categories. The above issues make the model more inclined to classify utterances with other emotional analogies as neutral emotions when classifying.
	
Using the T-SNE visualization tool, we examined the learned characteristics of the model, as shown in Fig. \ref{fig_8}. {Compare MGLRA (Fig. \ref{fig_8}-d) with original data (Fig. \ref{fig_8}-a), LSTM model (Fig. \ref{fig_8}-b) and DialogueRNN (Fig. \ref{fig_8}-c). MGLRA exhibits a more concentrated feature distribution, indicating superior clustering ability. In contrast, the feature distribution of the LSTM model, especially the feature distribution in the upper right corner, blurs the distinction between Angry, Sad, and Neutral emotions to a certain extent, causing the overall distribution to be more dispersed. Likewise, DialogueRNN has difficulty distinguishing between Happy and Neutral emotions and merging their data distributions.} MGLRA (Fig. \ref{fig_8}-d) demonstrates an impressive ability to cluster emotions tightly, and its emotion classification is more precise and effective.
	
	\begin{figure*}[htbp]
		\centering
		\begin{minipage}[b]{0.24\linewidth}
			\centering
			\includegraphics[width=\linewidth]{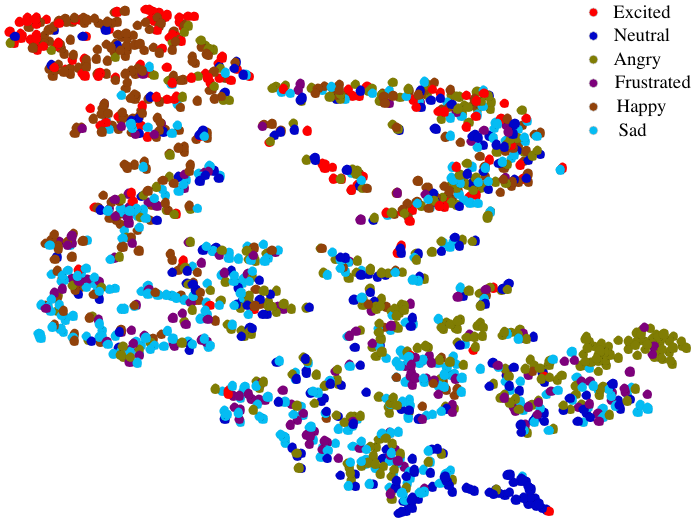}
			\scriptsize
			\\  (a) Origin data distribution

		\end{minipage}\hfill
		\begin{minipage}[b]{0.24\linewidth}
			\centering
			\includegraphics[width=\linewidth]{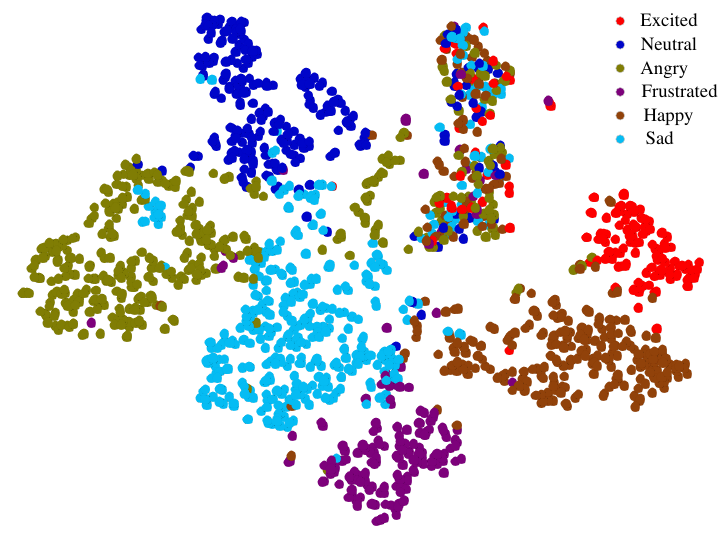}
			\scriptsize
			\\ (b) LSTM
			\label{fig:image2}
		\end{minipage}\hfill
		\begin{minipage}[b]{0.24\linewidth}
			\centering
			\includegraphics[width=\linewidth]{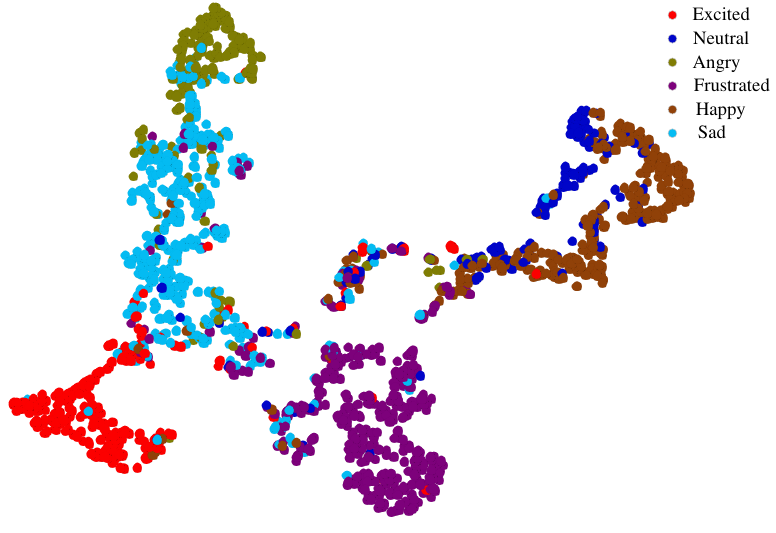}
			\scriptsize
			\\(c) DialogueRNN
			\label{fig:image3}
		\end{minipage}\hfill
		\begin{minipage}[b]{0.24\linewidth}
			\centering
			\includegraphics[width=\linewidth]{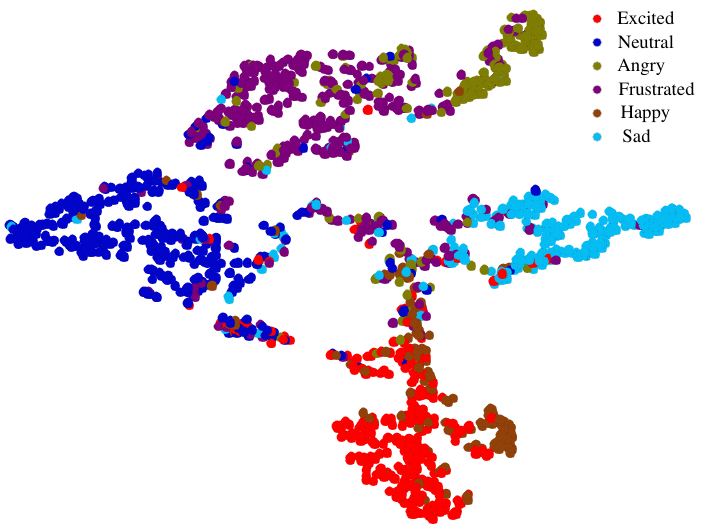}
			\scriptsize
			\\(d) MGLRA
			\label{fig:image4}
		\end{minipage}
		\captionsetup{labelformat=empty}
		\caption{Visualization of the feature embedding space on the IEMCOAP dataset using T-SNE. In this visualization, each emotion category is represented by a distinct color. (a) displays the original modal distribution of the IEMOCAP dataset, while (b), (c), and (d) demonstrate the modal distribution of the dataset after applying LSTM, DialogueRNN and MGLRA. }
		\label{fig_7}
	\end{figure*}

\subsection{{Comparison of time and total number of parameters}}
\begin{table}[htbp]
	\centering
    \caption{Performance evaluation of various models on IEMOCAP: comparison of time, number of parameters, {and performance.}}
	\renewcommand\arraystretch{1.5}
	\setlength{\tabcolsep}{2pt} 
	
	\begin{tabular}{c|cc|cc|c}
		\hline
		Dataset Modal & \multicolumn{2}{c|}{Number of Parameters (MB)} & \multicolumn{2}{c|}{Time (s)} & {Acc. F1} \\ \hline
		DialogueGCN & \multicolumn{2}{c|}{16.85} & \multicolumn{2}{c|}{78.3} & {65.1 64.2} \\
		LR-GCN	 & \multicolumn{2}{c|}{15.77} & \multicolumn{2}{c|}{76.1} & {68.4 68.2} \\
        GraphCFC & \multicolumn{2}{c|}{41.45} & \multicolumn{2}{c|}{103.7} & {69.1 68.5} \\
        RGAT	 & \multicolumn{2}{c|}{16.89} & \multicolumn{2}{c|}{79.6} & {65.0 65.2} \\
		DER-GCN	 & \multicolumn{2}{c|}{78.59} & \multicolumn{2}{c|}{98.4} & {69.7 69.4} \\
		\hline
		MGLRA (no mask) & \multicolumn{2}{c|}{17.45} & \multicolumn{2}{c|}{82.1} & {70.7 69.5} \\
		MGLRA (mask) & \multicolumn{2}{c|}{\textbf{13.21}} & \multicolumn{2}{c|}{\textbf{55.5}} & {\textbf{71.3 70.1}} \\ \hline
	\end{tabular}
	\label{tab-time}
\end{table}

{In Table \ref{tab-time}, we perform a comprehensive performance evaluation of the benchmark models on the IEMOCAP dataset, focusing on the number of parameters required for each model (megabytes) and the computation time (seconds). {To make the comparison fair, we only select GCN-based baseline methods for comparison, such as DialogueGCN \cite{7}, LR-GCN \cite{17}, RGAT \cite{ishiwatari2020relation}, GraphCFC \cite{li2023graphcfc}, and DER-GCN \cite{ai2024gcn}.} Specifically, MGLRA with the mask mechanism achieves the lowest memory footprint, with a parameter of only 13.21 MB, and the fastest training time of 55.5 seconds. This efficiency is attributed to our innovative masking mechanism, which effectively reduces redundancy in computing and memory usage without affecting the ability to capture relevant features from the data. This improves the efficiency of the model. {In addition, from a performance perspective, using masks not only improves the efficiency of the model but also promotes the effect of the model. Acc and F1 are both improved by 0.6, respectively.} The effectiveness of our method in handling complex conversation-based emotion recognition tasks makes it a highly competitive model in terms of efficiency and performance.}

\subsection{{Importance of the Modalities}}
{To assess the significance of text, vision, and audio modalities in emotion recognition, we conducted experiments on the IEMOCAP and MELD datasets. Our objective was to evaluate the impact of unimodal, bimodal, and multimodal features on model performance. These experiments are detailed in Table \ref{tab4}. Considering the datasets' data imbalance issue, we selected Weighted F1 Score (WF1) as our primary metric for its balanced consideration of precision and recall rates. Weighted Accuracy (WA) is a secondary metric. Our findings reveal that text-based features outperform those derived from audio and vision among the unimodal approaches. Specifically, text modality achieved WA scores of 63.1\% and 62.7\% and WF1 scores of 63.9\% and 61.8\% on the IEMOCAP and MELD datasets, respectively. These results underscore the text modality's pivotal role in our model's emotion recognition capability. Audio features ranked second in effectiveness, with WA scores of 61.3\% and 62.0\% and WF1 scores of 61.5\% and 61.2\% on the respective datasets. Vision features demonstrated the minor utility for emotion recognition, evidenced by WA scores of 57.7\% and 60.4\% and WF1 scores of 57.2\% and 60.5\% on IEMOCAP and MELD, respectively. This suggests challenges in extracting valuable emotional cues from vision data. Overall, our analysis indicates that text features introduce the slightest noise, thus facilitating more effective learning of dynamic feature representations by the model.}

\begin{table}[htbp]
	\centering
	\renewcommand\arraystretch{1.5}
	\setlength{\tabcolsep}{12pt} 
	\caption{THE EFFECT OF MGLRA ON TWO DATASETS USING UNIMODAL FEATURES AND MULTIMODAL FEATURES, RESPECTIVELY. T, V, AND A REPRESENT TEXT, VISION, AND AUDIO MODALITY FEATURES.}
	\begin{tabular}{c|cccccc}
		\hline
		\multirow{2}{*}{Modality} & \multicolumn{4}{c}{\textbf{IEMOCAP}} & \multicolumn{2}{c}{\textbf{MELD}} \\ \cline{2-7}
		& \multicolumn{2}{c}{WA} & \multicolumn{2}{c}{WF1} & \multicolumn{1}{c}{WA} & WF1 \\ \hline
		T & \multicolumn{2}{c}{63.1} & \multicolumn{2}{c}{63.9} & \multicolumn{1}{c}{62.7} & 61.8 \\
		A & \multicolumn{2}{c}{61.3} & \multicolumn{2}{c}{61.5} & \multicolumn{1}{c}{62.0} & 61.2 \\
		V & \multicolumn{2}{c}{57.7} & \multicolumn{2}{c}{57.2} & \multicolumn{1}{c}{60.4} & 60.5 \\
		T+A & \multicolumn{2}{c}{66.8} & \multicolumn{2}{c}{65.7} & \multicolumn{1}{c}{63.7} & 62.5 \\
		T+V & \multicolumn{2}{c}{65.4} & \multicolumn{2}{c}{64.7} & \multicolumn{1}{c}{63.2} & 63.3 \\
		V+A & \multicolumn{2}{c}{62.2} & \multicolumn{2}{c}{61.9} & \multicolumn{1}{c}{60.2} & 59.7 \\
		T+A+V & \multicolumn{2}{c}{\textbf{71.3}} & \multicolumn{2}{c}{\textbf{70.1}} &\multicolumn{1}{c}{\textbf{66.4}} & \textbf{64.9}  \\ \hline
	\end{tabular}
	\label{tab4}
\end{table}

{The comparison between bi-modal and single-modal approaches shows a notable enhancement in performance, with WA seeing an increase of 2\% to 9\% and WF1 improving by 2\% to 10\%. This improvement underscores that the context of conversations and the variations in audio-useful signals and visual cues from facial expressions influence emotional recognition. Bi-modal features significantly enhance the model's ability to predict emotions by integrating two distinct modalities. Among the bi-modal combinations, the fusion of text and audio modalities emerged as the most effective for emotion prediction, achieving WA scores of 66.8\% and 63.7\% and WF1 scores of 65.7\% and 62.5\%, respectively. The combination of text and video modalities ranked second in performance, with WA scores of 65.4\% and 63.2\% and WF1 scores of 64.7\% and 63.3\%, respectively. The fusion of audio and video modalities was the least effective, resulting in the lowest emotion prediction performance, with WA scores of 62.2\% and 60.2\% and WF1 scores of 61.9\% and 59.7\%, respectively. These findings highlight the strategic advantage of combining modalities for improved emotion prediction accuracy.}

{Integrating the modal features of text, vision, and audio leads to superior emotion prediction performance, with multimodal features outperforming those of single-modal and bi-modal configurations. This improvement suggests that the model leverages more than just the semantic content of dialogues. It also capitalizes on vision and audio cues to enrich the emotional feature vectors' representational capacity.}

\subsection{Ablation Study}
In this section, we present the experimental results of devoiding each part in MGLRA and analyze their performance on the IEMOCAP dataset to see their impact on performance. The corresponding results are shown in Table \ref{tab5}.
\begin{table}[]
	\centering
	\renewcommand\arraystretch{1.5}
	\caption{Ablation experimental performance of MGLRA on the IEMOCAP dataset with different components of MGLRA.}
	\begin{tabular}{c|cccc|cc}
		\hline
		\multirow{2}{*}{Row Number} & \multicolumn{4}{c|}{Components}                                                   & \multicolumn{2}{c}{Performance}                    \\ \cline{2-7}
		& \multicolumn{1}{c|}{GAF} & \multicolumn{1}{c|}{MRFA} & \multicolumn{1}{c|}{MHA} & MG & \multicolumn{1}{c|}{Acc.}          & F1            \\ \hline
		1                           & \multicolumn{1}{c|}{-}  & \multicolumn{1}{c|}{-}  & \multicolumn{1}{c|}{-}   & -  & \multicolumn{1}{c|}{63.5}          & 63.3          \\ \hline
		2                           & \multicolumn{1}{c|}{\textsurd}  & \multicolumn{1}{c|}{-}  & \multicolumn{1}{c|}{-}   & -  & \multicolumn{1}{c|}{65.7}          & 65.1          \\ \hline
		3                           & \multicolumn{1}{c|}{\textsurd}  & \multicolumn{1}{c|}{\textsurd}  & \multicolumn{1}{c|}{-}   & -  & \multicolumn{1}{c|}{68.3}          & 67.6          \\ \hline
		4                           & \multicolumn{1}{c|}{\textsurd}  & \multicolumn{1}{c|}{\textsurd}  & \multicolumn{1}{c|}{\textsurd}   & -  & \multicolumn{1}{c|}{70.7}          & 69.5          \\ \hline
		5                           & \multicolumn{1}{c|}{\textsurd}  & \multicolumn{1}{c|}{\textsurd}  & \multicolumn{1}{c|}{\textsurd}   & \textsurd  & \multicolumn{1}{c|}{\textbf{71.3}} & \textbf{70.1} \\ \hline
	\end{tabular}
	\label{tab5}
\end{table}
	
\begin{itemize}
\item[$\bullet$] Comparing the first and second rows shows the impact of {the graph} attention filtering mechanism (GAF) on performance. Compared with emotion classification using only LSTM's discourse representation, graph filtering can improve accuracy by 2.2\%  because the graph filtering mechanism considers the impact of noise generated in the dialogue.
\item[$\bullet$] Comparing the second and third rows shows the impact of memory-based recursive feature alignment (MRFA) on enhancing semantic features. We can see that using loop alignment improves accuracy by 2.6\%  compared to graph filtering alone, which shows that it effectively aligns semantic features from multiple modalities to facilitate late fusion performance.
\item[$\bullet$] The results in the third and third rows show the effect of adopting multi-head attention (MHA) on performance. Compared with the recurrent alignment without multi-head attention, the accuracy is improved by 2.4\%, which shows that the multi-head attention mechanism effectively captures the correlation between utterances from multiple modalities during the alignment process.
\item[$\bullet$] The impact of adopting multimodal fusion with masked GCN (MG) on emotion prediction is shown in the fourth and fifth rows. The use of masked GCN not only brought a 0.6\% increase in accuracy but also effectively reduced the model's running memory and time, facilitating the deployment and operation of our model on low-performance machines.	
\end{itemize}
	
When we adopt GAF, MRFA, MHA, and MG at the same time, MGLRA achieves the best performance. These combinations constitute our final model, and the rationality of our model is proved by experiments.

\subsection{Analysis on Parameters}
\begin{figure}[htbp]   
	
	\includegraphics[width=\linewidth,scale=1.00]{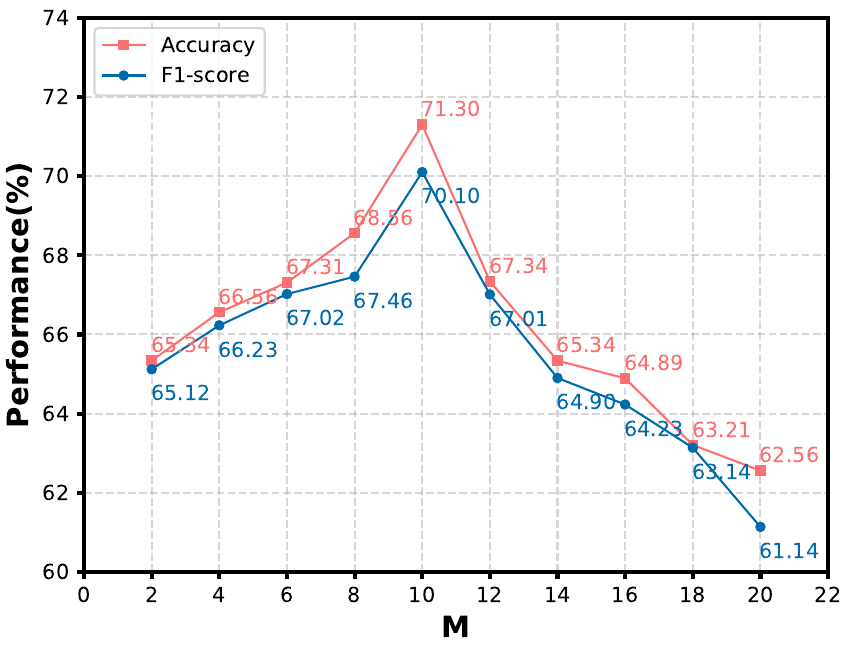}
	\caption{MGLRA experimental results of the number of heads in multi-head attention
		Graphs on the IEMOCAP dataset.}
	\label{fig_8}
\end{figure}

\begin{figure}[htbp]   
	
	\includegraphics[width=\linewidth,scale=1.00]{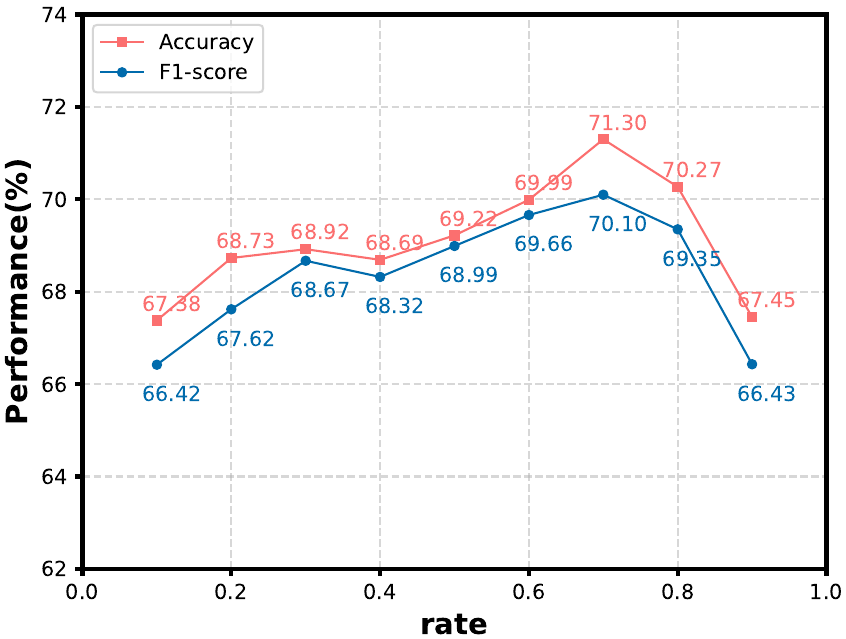}
	\caption{Experimental results of MGLRA with different masked rate on the IEMOCAP dataset.}
	\label{fig_9}
\end{figure}

In this section, we analyze the impact of the head number $M$ in multi-head attention and the masked rate in masked GCN on the model. The relevant experimental results are shown in Fig. \ref{fig_8} and Fig. \ref{fig_9}, respectively.
	
$M$ in the multi-head attention mechanism is a vital hyperparameter for the experimental results because it relates to the depth of potential relationship exploration in the alignment process. The masked rate in masked GCN is another hyperparameter that determines the performance of MGLRA. It is directly related to memory usage, the time required for model deployment, and the propagation depth of features in the model. We choose $M$ and masked rate through comparative experiments. Specifically, we initially choose the most suitable $M$ from $\{2,4,6,8,10,12,14,16,20\}$ and choose the most appropriate masked rate from $\{0.1, 0.2, 0.3, 0.4, 0.5, 0.6, 0.7, 0.8, 0.9\}$ . In the experiment, we set $M$ to 10 to get the experimental results about the asking rate and put the masked rate to 0.7 to get the testing results about $M$.
	
The experimental results in Fig. \ref{fig_8} prove that MGLRA performance is proportional to $M$ in the range of 2 to 10. When $M=10$, MGLRA achieves the best performance, and the accuracy is 71.3\%. However, when $M>10$, the performance of MGLRA starts to drop significantly. It can be seen from the experiments that when exploring more potential relations, multi-head attention cannot provide useful semantic information, and brings more redundant information, resulting in a decline in model performance.
	
Furthermore, increasing the masked rate from 0.1 to 0.7 gradually improves performance. As shown in Fig. \ref{fig_9}, we observed that MGLRA achieves the best performance when rate $=$ 0.7; at this time,  accuracy $=$ 71.3\%, and F1-score $=$ 70.10\%, which shows that shielding more nodes will only lead to the loss of much crucial emotional information. This process reduces the richness of emotional communication, which directly limits the performance of MGLRA.

\section{Conclusion}
\label{sec:conclusion}
In this study, we propose a recurrent alignment method to enhance the features of each modality and make the semantic gap between modalities more minor, making up for some shortcomings of current SOTA methods such as GraphCFC. Simultaneously, we incorporate a directed graph-based masked GCN to enhance the model's generalization ability and reduce memory usage. Our proposed MGLRA approach consistently surpasses existing SOTA models through experimental evaluation on two public datasets. These results demonstrate the effectiveness of our work in aligning semantic information between modalities and enhancing self-representation features.
Many recent studies have shown that MERC still faces challenges, such as semantic gaps between modalities and many noises within modalities. In future work, we plan to optimize multimodal fusion and semantic information alignment methods and evaluate whether masked GCN applies to other multimodal tasks.

\section*{Acknowledgments}
{The authors deepest gratitude goes to the anonymous reviewers and AE for their careful work and thoughtful suggestions that have helped improve this paper substantially.}

\bibliographystyle{IEEEtran}
\bibliography{IEEEabrv,reference.bib}

\newpage
	
	
\vspace{11pt}
	
	
	\vspace{11pt}
	
	
	\vfill
	
\end{document}